\documentclass[10pt,twocolumn,letterpaper]{article}
\usepackage[pagebackref=true,breaklinks=true,letterpaper=true,colorlinks,bookmarks=false]{hyperref}
\usepackage{wacv}
\usepackage{times}
\usepackage{epsfig}
\usepackage{graphicx}
\usepackage{amsmath}
\usepackage{amssymb}
\usepackage[noadjust]{cite}
\usepackage{multirow}
\usepackage{algorithm}
\usepackage{algorithmicx}
\usepackage{epstopdf}
\usepackage{enumitem}
\usepackage[noend]{algpseudocode}

\graphicspath{{./Figures/}}

\usepackage{url}

\wacvfinalcopy 

\def\wacvPaperID{84} 

\ifwacvfinal\fi
\begin{document}


\title{Discovering Useful Parts for Pose Estimation in Sparsely Annotated Datasets}


\author{Mikhail Breslav$^1$, Tyson L. Hedrick$^2$, Stan Sclaroff$^1$, and Margrit Betke$^1$\\
$^1$Department of Computer Science and $^2$Department of Biology \\
$^1$Boston University and $^2$University of North Carolina \\
{\tt\small breslav@bu.edu, thedrick@bio.unc.edu, sclaroff@bu.edu, betke@bu.edu}}

\maketitle
\ifwacvfinal\thispagestyle{empty}\fi

\begin{abstract}
Our work introduces a novel way to increase pose estimation accuracy by discovering parts from unannotated regions of training images. Discovered parts are used to generate more accurate appearance likelihoods for traditional part-based models like Pictorial Structures \cite{Felzenszwalb2005} and its derivatives. Our experiments on images of a hawkmoth in flight show that our proposed approach significantly improves over existing work \cite{Ortega-Jimenez2014} for this application, while also being more generally applicable. Our proposed approach localizes landmarks at least twice as accurately as a baseline based on a Mixture of Pictorial Structures (MPS) model. Our unique High-Resolution Moth Flight (HRMF) dataset is made publicly available with annotations.  
\end{abstract}

\section{Introduction}
Researchers are actively studying flying animals to better understand their behaviors and flight characteristics. For example, researchers study the group behavior and obstacle avoidance abilities of bats \cite{Breslav2012,Breslav2014,Kong2013,Theriault2010}, the maneuverability of cliff swallows involved in a chase \cite{Shelton2014}, and the flight performance of hawkmoths under varying wind tunnel conditions \cite{Ortega-Jimenez2013,Ortega-Jimenez2014}. Enabling this research to take place are camera systems, which have been essential for the observation of flying animals in both lab conditions and natural habitats \cite{Breslav2012,Breslav2014,Ortega-Jimenez2013,Ortega-Jimenez2014,Shelton2014,Theriault2010}. Analyses of the datasets captured by these camera systems have increasingly been assisted by computer vision algorithms, allowing researchers to save time and labor on tasks that algorithms can do automatically with sufficient accuracy.  

Our research is inspired by the question of how hawkmoths ({\it Manduca sexta}) fly in varying wind conditions, a problem recently studied by Ortega-Jimenez et al. \hspace{-0.5cm} \cite{Ortega-Jimenez2013}. In their work, hawkmoths were placed into a wind tunnel where their flight was captured using multiple high-resolution, high frame-rate cameras. To analyze the flight sequences of hawkmoths, computer vision was used. First key body landmarks were localized across multiple camera views and time. Secondly, 3D positions of these landmarks were reconstructed across time. While Ortega-Jimenez et al. \cite{Ortega-Jimenez2013} obtained interesting results, their approach to landmark localization only works on datasets where the hawkmoth is observed from a particular view point, thus limiting the general applicability of their approach. By contrast, we propose an approach for landmark localization that does not place any restrictions on the view point. For the rest of the paper we will use the term landmark localization interchangeably with `pose estimation', as the pose of a hawkmoth, and animals in general, can be specified by the position (localization) of key landmarks.

\begin{figure}[t]
\begin{center}
   \includegraphics[width=0.5\linewidth]{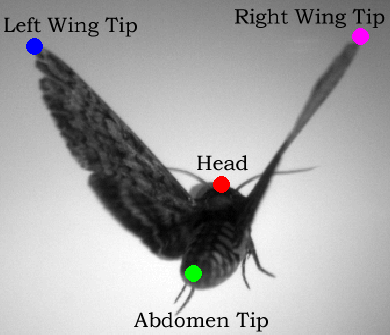} \\
\end{center}
   \caption{A Hawkmoth ({\it Manduca sexta}) is viewed from behind. Four key body landmarks sufficient for describing the pose of the moth are labeled with text alongside a colored circle at the location of the landmark.}
\label{fig:labeledmoth}
\vspace{-0.3cm}
\end{figure}

In the field of computer vision, pose estimation is a fundamental research problem that has received a lot of attention.  Among the large body of works that exist, part-based models in particular have shown great success in both 2D and 3D human pose estimation \cite{Amin2013,Andriluka2012,Bourdev2009,Bourdev2010,Burenius2013,Felzenszwalb2005,Fischler1973,Gkioxari2014,Hernandez2014,Kazemi2013,Pishchulin2013,Sun2011,Wang2011,Yang2011,Zhu2012}. Part-based models have a common approach of modeling an object by a collection of parts. The definition of what a part is varies, but common to all of the mentioned approaches, the representation of a part is learned from annotations provided with training images.

We argue that the complete dependence of part-based models on annotations is a weakness, especially limiting in applications where training data is sparsely annotated. Consider the problem of localizing the positions of four landmarks of interest, shown in Figure \ref{fig:labeledmoth}, on a hawkmoth test image. Assume training images are given with those same landmarks annotated. Part-based models would do their job of modeling parts based on annotations, while regions of the training images without annotations, including most of the wings, abdomen, and antennae, risk being thrown away. Thrown away regions may contain parts which are helpful for localizing the landmarks of interest. We hypothesize that augmenting traditional part-based models with parts discovered from the unannotated regions of training images can improve the localization accuracy of landmarks of interest, especially in sparsely annotated datasets. \\

\noindent We now summarize the main contributions of our work. 

\begin{enumerate}[leftmargin=*,itemsep=1pt]
\item We propose a novel approach to pose estimation in sparsely annotated datasets. Our approach augments traditional part-based models with useful information derived from parts that are discovered automatically from unannotated regions of training images. 
\item We demonstrate experimentally that our approach leads to better pose estimation accuracy compared with a baseline representative of traditional part-based models. 
\item We show that our approach is well suited for the problem of hawkmoth pose estimation and is more general and more accurate than the recent work by Ortega-Jimenez et al. \cite{Ortega-Jimenez2014}.
\item We introduce the HRMF (High-Resolution Moth Flight) dataset, which as far as we know will be the first high-resolution, high frame-rate, video dataset capturing detailed flight of a flying animal that is made publicly available with part annotations and segmentations. 

\end{enumerate}

%

\section{Related Work}
\label{sec:related}

Our work lies at the intersection of natural science research on flying animals, like hawkmoths, and mainstream computer vision research. Here we give our work context with respect to both communities.

\subsection{Natural Science Community}
In the natural science community, researchers have taken advantage of pose estimation algorithms to study the behavior and flight characteristics of flying animals such as: bats \cite{Bergou2011,Hubel2012}, birds \cite{Shelton2014,Tobalske2007}, flies \cite{Fontaine2009,Ristroph2009}, and moths \cite{Ortega-Jimenez2013,Ortega-Jimenez2014}. 

One common approach for estimating the 3D pose of flying animals in laboratory conditions relies on the placement of physical markers like tape on key landmarks across the animal's body \cite{Bergou2011,Hubel2012,Tobalske2007}. These markers, which are visible in recorded video datasets, are localized in multiple views either manually or automatically. Landmark locations across views are then reconstructed in 3D yielding pose estimates. In Shelton et al. \cite{Shelton2014}, cliff swallows were observed in a natural environment precluding the use of markers. Manual annotations were relied on to localize landmarks in the image data, which were subsequently used to estimate pose.

More automated approaches for pose estimation of flying animals include \cite{Breslav2014,Fontaine2009,Ortega-Jimenez2013,Ortega-Jimenez2014,Ristroph2009}. In Ristroph et al. \cite{Ristroph2009} multiple views of a fruit fly are segmented and then back projected for visual hull reconstruction. Reconstructed voxels are clustered into groups corresponding to different body parts. A final 3D pose estimate is obtained by computing the position and orientation of part clusters. Fontaine et al. \cite{Fontaine2009} track the 3D pose of a fly over time by registering a 3D articulated graphics model with segmented image data. A similar approach was used by us for 3D pose estimation of bats \cite{Breslav2014}. 

In Ortega-Jimenez et al. \cite{Ortega-Jimenez2013,Ortega-Jimenez2014}, work we aim to improve over, a hawkmoth is segmented in multiple camera views and then various heuristics are used to localize the image location of the head, abdomen tip, left wing tip, and right wing tip (Figure \ref{fig:labeledmoth}). Specifically, the left and right wing tips were localized in one of the camera views for frames where the moth was in a particular phase of its wingbeat cycle. The head and abdomen tip were localized by removing the wings from the segmented moth using temporal information and then using morphological operations to remove the antenna and proboscis of the moth. Extrema along the boundary of the remaining connected component were then classified as the head and abdomen tip. To localize landmarks across all camera views epipolar geometry was leveraged.

%

\subsection{Computer Vision Community}

In the context of computer vision, our approach to pose estimation combines ideas from established part-based models \cite{Bourdev2009,Felzenszwalb2005,Fischler1973}, with recent works on unsupervised or weakly supervised part discovery \cite{Doersch2013,Juneja2013,Singh2012}. 

One established part-based model is {\it pictorial structures} (PS) \cite{Felzenszwalb2005,Fischler1973} which 
continues to be the foundation for many 2D and 3D human pose estimation works \cite{Amin2013,Andriluka2012,Burenius2013,Hernandez2014,Kazemi2013,Pishchulin2013,Yang2011,Zhu2012}. PS is a model that integrates the appearance of individual parts (unary terms) with preferred spatial relationships between parts (pairwise terms). Many PS-based works have a one-to-one mapping between parts in the model and annotations provided with the training images \cite{Amin2013,Andriluka2012,Burenius2013,Kazemi2013,Yang2011,Zhu2012}. As a result, these models ignore regions of the training images that are unannotated. If unannotated regions contain useful parts then these models cannot leverage them. In contrast, our work augments traditional PS-based models with useful parts discovered from unannotated regions.

One exception to the reliance of part-based models on part annotations is the Deformable Part Models (DPM) work \cite{Felzenszwalb2010} which learns parts with only bounding-box level supervision. While DPMs have shown success in object detection, they are not well suited for pose estimation applications where specific landmarks need to be localized. There is no guarantee that parts learned by a DPM will correspond to landmarks that need to be localized.

Another established part-based model is the work of Bourdev et al. \cite{Bourdev2009} who introduce {\it poselets}. Poselets, can be thought of as mid-level parts that capture common configurations of low-level parts. Specifically, a single poselet (part) is defined by a set of visually similar image patches that contain similar configurations of annotations. This broader definition of part has proven to be useful for pose estimation as seen in the success of recent works \cite{Bourdev2010,Gkioxari2014,Hernandez2014,Pishchulin2013,Wang2011}. Unfortunately, like traditional parts, poselets are dependent on annotations and cannot capture parts from regions of training images that neither contain nor are near annotations. 

Looking beyond pose estimation, there have been recent works on unsupervised and weakly supervised part discovery \cite{Doersch2013,Juneja2013,Singh2012}. These works showed the utility of the parts they discovered by using them as feature representations of scenes for supervised scene classification. Our work takes inspiration from these methods and uses a simpler part discovery approach for the problem of pose estimation.

\section{Methods}
\label{sec:methods}

\begin{figure*}[t]
\begin{center}
   \includegraphics[width=1.0\linewidth]{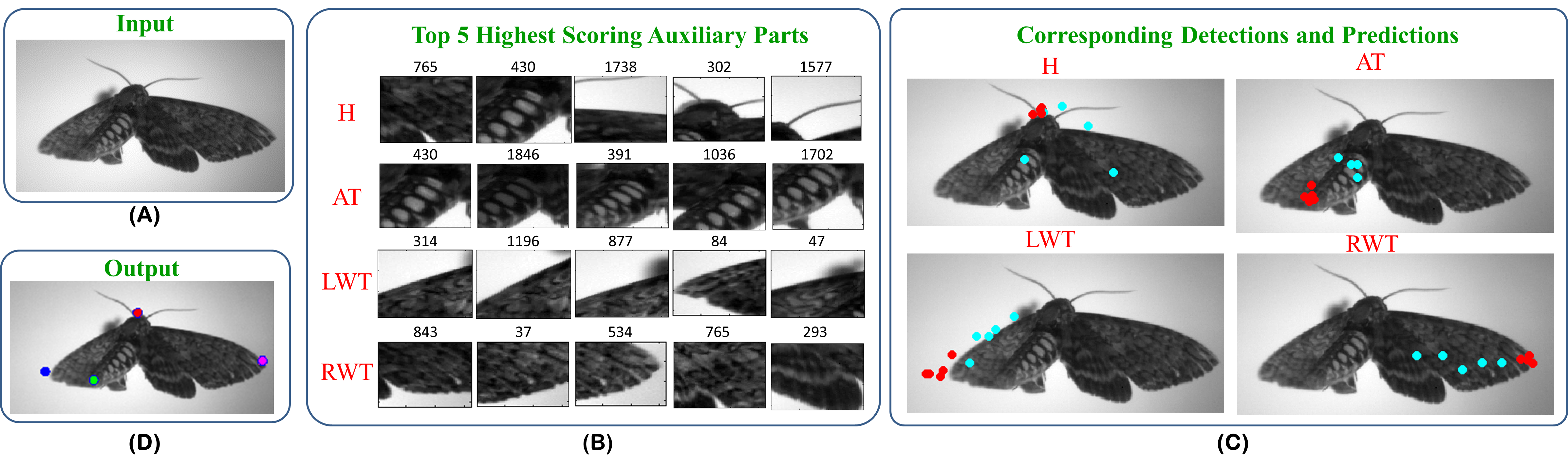} \\
\end{center}
	\vspace{-0.3cm}
   \caption{Illustrating example of how auxiliary parts are leveraged: (a) Input test image. (b) For each semantic part (\textbf{H}ead, \textbf{A}bdomen \textbf{T}ip, \textbf{L}eft \textbf{W}ing \textbf{T}ip, and \textbf{R}ight \textbf{W}ing \textbf{T}ip) we show the top 5 highest scoring auxiliary parts. Each auxiliary part is labeled by an index and represented by one example patch. (c) For each semantic part we show where the top 5 auxiliary parts were detected on the test image (cyan circles) and where their predictions or votes went (red circles). Notice that the predictions are visually close to the location of the semantic part. Also note that the votes are weighted but for simplicity we do not reflect that in our visualization. (d) After integrating votes from {\it all} auxiliary parts with the Mixture of Pictorial Structures model we obtain the final pose estimate (part localizations) shown. }
\label{fig:proposed_overview}
\vspace{-0.3cm}
\end{figure*}


We begin this section by specifying our research problem and giving an overview of the proposed solution. \\

\noindent \textbf{Problem:} Given a test image of an object or animal of interest, estimate the image locations $L = \{L_1,L_2,\cdots,L_n\}$ of a set of predetermined landmarks $S = \{S_1,S_2,\cdots,S_n\}$. The estimated locations $L_i = (x_i,y_i)$ of landmarks $S_i$ should be as close as possible to ground truth locations $L^*_i$. We assume that a training set is available as input, where images containing the object or animal of interest have the image locations of landmarks $S$ annotated. 

While this problem formulation is quite general, our application targets the study of flying animals like hawkmoths. As a result, the landmarks to be localized are parts of the animal's body that are meaningful for the analysis of flight. Therefore, for the remainder of the paper we will refer to the landmarks to be localized as semantic parts. \\

\noindent \textbf{Proposed Solution (Overview):} Our proposed solution takes a representative part-based model (Section \ref{sec:mops}) and modifies the part appearance likelihood terms so that semantic parts can be localized with higher accuracy. Part appearance likelihoods are improved by incorporating useful information obtained from parts discovered automatically from {\it unannotated} regions of training images (Section \ref{sec:dap}). \\ 

\noindent \textbf{Sparsely Annotated Datasets:} Our work targets datasets that are sparsely annotated like the hawkmoth dataset, a sample of which is shown in Figure \ref{fig:labeledmoth}. Formally, a dataset is sparsely annotated when there exist one ore more unannotated `parts' present in training images, whose detection would be predictive of the location of one or more semantic parts. This loose definition means many datasets are sparsely annotated. In the case that a dataset contains annotations for all useful `parts' and is therefore not sparsely annotated, our approach will not have any useful parts to discover and it will default to a standard part-based model.


\subsection{Mixture of Pictorial Structures}
\label{sec:mops}

The basis of our approach is a mixture of pictorial structures (MPS) model. A pictorial structures (PS) model \cite{Felzenszwalb2005} represents an object or animal by a collection of parts whose relationships are modeled with a tree structured graph. Mathematically a PS model can be written as 

\vspace{-.2cm}
\begin{equation} p(L|I) \propto p(I|L) \; p(L) \end{equation} where $p(L)$ is a prior distribution on the locations of $n$ parts, and $p(I|L)$ is the appearance likelihood term describing how well image evidence $I$ agrees with part configuration (localization) $L$. The posterior distribution $p(L|I)$ describes the most probable part configuration(s) given the image evidence. The spatial relationship between parts is encoded by a tree structured graph with vertices $V$ and edges $E$. If $(i,j)$ is an edge in $E$ then there is a preferred spatial relationship between parts $i$ and $j$ in the model. Using this information, along with the assumption that parts do not overlap in the image evidence, the posterior can be rewritten as:\vspace{-.2cm} \begin{equation} p(L|I) \propto \bigg[ \prod_{i=1}^{n}{p(I|L_i)} \prod_{(i,j)\in E}{p(L_i,L_j)} \bigg] \end{equation} \vspace{-.2cm}

\noindent PS models can also be used as individual components of a mixture model, which is an effective way to capture variation in the appearance and relationship of parts due to changes in pose \cite{Amin2013,Felzenszwalb2010,Johnson2010,Zhu2012}. A mixture of PS can be written as: \vspace{-.2cm} \begin{equation} p(L^k|I) \propto \bigg[ \prod_{i=1}^{n}{p(I|L^k_i)} \prod_{(i,j) \in E^k}{p(L^k_i,L^k_j)} \bigg] \end{equation} where $p(L^k|I)$ is the posterior given by the $k^{th}$ pictorial structure in the mixture, $k \in \{1,\cdots,m\}$. The objective for the MPS model can then be stated as finding the most probable part configuration among all PS components: \begin{equation} \arg\max_{L^k} \quad p(L^k|I) \end{equation} 

\noindent Further design and implementation details of the MPS model are provided in Section \ref{sec:baseline}.

%
%
%
%

%


\subsection{Discovering Auxiliary Parts}
\label{sec:dap}

The MPS model described in Section \ref{sec:mops} is a good starting point for localizing semantic parts. However, there are potentially useful parts in the training images that are not annotated, which a MPS model (like most other part-based models) cannot make use of. Instead of letting these potentially useful parts go to waste we think of them as auxiliary parts that can be discovered and incorporated into a MPS model. We say an auxiliary part is useful if its presence in an image can be used to predict where one or more semantic parts are located. Auxiliary parts are not required to have semantic meaning.

To discover {\it useful} auxiliary parts, we first discover auxiliary parts and then determine which are useful (predictive) enough to keep. Discovery of auxiliary parts begins by an image patch generation step. All training images are segmented and image patches are extracted from the segmented regions. To avoid generating too many patches, no patches can be extracted too near to an already extracted patch. The large set of patches generated by this step is then represented by a feature and clustered into visually similar clusters. A single auxiliary part can then be thought of as a model of the appearance of patches belonging to a particular cluster. In Section \ref{sec:auxparts}, we detail our choices of features, as well as segmentation and clustering algorithms. 

Every patch in a cluster has associated with it the training image it came from, the image location it was extracted from, and the image locations of annotated semantic parts. Using this information, each auxiliary part (corresponding to some cluster $C$) can be evaluated by how well it predicts one or more semantic parts. Suppose cluster $C$ contains patches $\{P_k\}, k \in \{1,\cdots,K\}$, and each patch $P_k$ is associated with image location $L_k = (x_k,y_k)$, from which it was extracted. Also, let $S^k_i = (x^k_i,y^k_i) $ be the image location of the $i^{th}$ semantic part annotation in the same training image as patch $P_k$ was extracted from. Then, the disagreement $D_i(C)$ on the relative position of semantic part $i$ relative to patch center $L_k$ can be computed across all patches in a cluster by: \begin{equation} D_i(C) = \frac{1}{K} \sum_{k=1}^{K}{ ||(S^k_i - L_k) - \mu^k_i ||} \end{equation} with \begin{equation} \mu^k_i = \frac{1}{K} \; \sum_{k=1}^{K}{(S^k_i - L_k)} \end{equation}
The smaller the disagreement $D_i(C)$, the more cluster $C$ is in agreement on the relative location of semantic part $i$. If disagreement $D_i(C)$ is less than some chosen threshold $\tau_i$, then the auxiliary part modeling cluster $C$ is considered predictive of semantic part $i$. We obtain the set of useful auxiliary parts $A$ by keeping all auxiliary parts that are predictive of at least one semantic part.  $A = \{C : \exists \; i, \; D_i(C) <= \tau_i \}$

%

\subsection{Leveraging Auxiliary Parts}
\label{sec:leverage_aux_parts}
Once useful auxiliary parts are discovered, they are used to update the appearance likelihoods of semantic parts in the MPS model. Each auxiliary part (we now drop the adjective useful and consider it implied) is realized as a discriminative classifier learned from patches belonging to its cluster. For a given test image, all auxiliary part detectors are evaluated on the image, and those scoring higher than a threshold are allowed to give a weighted vote for the locations of semantic parts that they are predictive of. The weight of the vote corresponds to the output of the detector, with a larger vote indicating a more confident detection. The prediction an auxiliary part makes is computed by taking the location where the auxiliary part is detected and adding $\mu^k_i$, the average displacement of semantic part $i$ with respect to that part. 
After all votes (predictions) are in, new appearance likelihoods are obtained for each semantic part by a linear combination of the existing appearance likelihoods with the weighted votes. The new appearance likelihoods are used in the MPS model to obtain final pose estimates.

We illustrate how discovered auxiliary parts are leveraged on a test image in Figure \ref{fig:proposed_overview}. Specifically, for each semantic part we show exemplars from just the top 5 highest scoring auxiliary parts. These auxiliary parts are then detected in the test image and their votes for semantic parts are recorded. The votes obtained for each semantic part can be thought of as a new response map or appearance likelihood which is integrated with an existing MPS model. The evaluation of the updated MPS model on the test image yields the final pose estimate shown in Figure \ref{fig:proposed_overview}d. Note the top 5 auxiliary parts are shown for ease of visualization but many more are used to obtain the output.


\section{System Design and Implementation}
The methods described in Section \ref{sec:methods} explain the underlying approach that we apply to hawkmoth pose estimation. 
However, the specific design and implementation of these methods is heavily influenced by the hawkmoth dataset, so in this section we first introduce the hawkmoth dataset and then discuss implementation design and details.

\subsection{Dataset}
\label{sec:dataset}


For our experiments we use a hawkmoth dataset from Ortega-Jimenez et al. \cite{Ortega-Jimenez2014} which captures an individual hawkmoth ({\it Manduca sexta}) hovering in a vortex chamber where the wind intensity is high. The hawkmoth dataset comes from a camera equipped with a 28 mm lens which records at 400 frames per second and has a resolution of 600 $\times$ 800 pixels. For all our experiments, we consider the semantic parts of a hawkmoth to be the left wing tip, right wing tip, abdomen tip, and head, the same parts that were identified as meaningful for 3D pose by biologists in \cite{Ortega-Jimenez2013,Ortega-Jimenez2014}. The experiments we perform in Section \ref{sec:experiments} evaluate how accurately different algorithms localize these four parts. 

To facilitate the evaluation of machine learning based algorithms on this dataset we annotate the image location of the four semantic parts in 421 images. The high-resolution hawkmoth image data along with part annotations and segmentations is being made publicly available\footnote{\url{http://www.cs.bu.edu/~betke/research/HRMF/}} to encourage more computer vision researchers to evaluate their approaches on this unique biology dataset.

\subsection{Design and Implementation Details}

The MPS model introduced in Section \ref{sec:mops} serves as a baseline algorithm in our work. In this section we give implementation details.

\subsubsection{Mixture of Pictorial Structures Baseline}
\label{sec:baseline}

The individual components of the mixture model are PS models with spatial terms learned from 2D pose clusters. Clustering of poses is done by first gathering annotated image locations of the four semantic parts across all training images. If a semantic part is occluded, an annotation is still provided using an educated guess. The 2D pose of a hawkmoth in a training image is then described by the 8 dimensional vector that contains $(x,y)$ annotations for the four semantic parts. These vectors are clustered using affinity propagation \cite{Frey2007} which requires an affinity (similarity) matrix as input. We define the distance between 2D poses $D(p_i,p_j)$ to be the Euclidean distance. The similarity is then computed by $S(p_i,p_j) = e^{-(\alpha \; D(p_i,p_j) )}$, where $\alpha$ is a scaling parameter. In our experiments we have 26 pose clusters so $m$, the number of PS in the mixture, is also 26.

We design the PS appearance terms to be shared across mixture components, since a part can appear similar across different poses. For a given part type ($k \in \{1,\cdots,K\}$), patches of size 64 $\times$ 64 centered on annotations of that part type are extracted from all training images and clustered into visually similar clusters. Patch appearances are represented with whitened HOG \cite{Hariharan2012} features (WHOG), using a cell size of 8 $\times$ 8 pixels. Affinity propagation \cite{Frey2007} is used to cluster patches of the same part-type. The similarity of two image patches is computed as the dot product of their respective WHOG features, $S(p_i,p_j) = f_i \cdot f_j$. The appearance of a part cluster is modeled by learning an LDA classifier on HOG \cite{Dalal2005} features, with the positive samples being patches in the cluster, and negative samples being all other patches as well as background patches. In our experiments the number of appearance terms we obtained for each part were: head: 32, abdomen tip: 27, left wing tip: 26, and right wing tip: 17. 

We determine which appearance terms are assigned to which visual clusters, the following logic is used:
Let $Y^k_l$, be the training image indices that are assigned to visual cluster $k$ for part-type $l \in \{1,\cdots,n\}$, and let $X_i$ be the training image indices that are assigned to 2D pose cluster $i$ (equivalently the $i^{th}$ PS component), $i \in \{1,\cdots,m\}$. Then, the part appearance represented by the visual cluster $Y^k_l$ is shared with the $i^{th}$ PS model if $Y^k_l$ and $X_i$ have a non-empty intersection.

When evaluating the resulting MPS model on a test image, all appearance terms assigned to a PS component are evaluated and the one scoring highest gives the score for the overall PS model. The evaluation of each PS model in the mixture on a test image (inference of the tree model) is done by dynamic programming, implemented using the generalized distance transform \cite{Felzenszwalb2004} for Gaussian spatial relationships.

 
%

\subsubsection{Auxiliary Parts}
\label{sec:auxparts}

\textbf{Segmentation and Features} \\
The first part of the patch generation step described in Section \ref{sec:dap} is segmentation of all training images. We perform segmentation of the hawkmoth by observing that most of a training image is brightly colored background. We use a histogram of image intensities to find the threshold where a fixed percentage of pixels are brighter than it. This threshold does a good job of segmenting most of the hawkmoth but tends to miss the antennae. To recover the antennae we add regions of the image that have a large gradient magnitude. Finally, connected components are computed and those that are too large or too small are removed. 

From the segmented region of the image, which corresponds to the hawkmoth, we uniformly sample patches of size 64 $\times$ 64 pixels, subject to the constraint that no patches are extracted from within 8 pixels of a previously extracted patch. Across all training images this results in approximately 36,000 patches. For each patch, dense SIFT \cite{Liu2011} is extracted and used to compute a bag of words (BOW) feature. The BOW dictionary is built using k-means on dense SIFT keypoints with $k = 500$. To preserve some spatial information the BOW feature is computed for a two level spatial pyramid. The resulting feature is the concatenation of a 500 dimensional histogram for the whole patch (first level), and 500 dimensional histograms for each of the four quadrants of the patch (second level). The total feature dimension is 2500. \\

\noindent \textbf{Clustering} \\
We cluster patches using a greedy strategy where clusters are formed one at a time until all patches have been considered. Algorithm \ref{alg:clustering} shows our algorithm in pseudocode. To form a cluster, first a seed patch $i$ is randomly selected from unclustered seeds $S$. Then from available patches $P$, the $k$ patches that are most similar to the seed are found using histogram intersection and used as the initial cluster $Q$. To ensure the cluster $Q$ is visually similar to and in agreement with the seed, several pruning steps are performed. 

The first pruning step involves computing an alignment energy (SIFT Flow \cite{Liu2011}) $E$ between all patches in $Q$ and the seed. If the alignment energy for any patch is above some threshold $\beta$, it is considered not visually similar enough to the seed and thus discarded from the cluster. The second pruning step involves computing the disagreement (Section \ref{sec:dap}) in semantic part prediction between each patch in $Q$ and the seed. If any patch disagrees with the seed by more than a threshold $\gamma$, the patch is considered to be an outlier not representing the same part in a similar pose as the seed and thus it is discarded. If the resulting cluster is too small (size less than $\alpha$) the seed patch is removed from $S$ and the process repeats. Otherwise, the remaining patches in $Q$ are joined with seed patch $i$ forming a cluster $C^*$ and added to clusterings $C$. Patches in $C^*$ are removed from $P$ and the process repeats until all seeds have been processed.

To make use of the clusters obtained by our clustering approach we model their appearance with discriminatively trained classifiers. Specifically, a cluster is modeled by extracting HOG features from its patches and then training an LDA classifier. The resulting LDA classifier can be thought of as a detector of an auxiliary part. Furthermore, each auxiliary part is associated with scores indicating how predictive it is of each of the semantic parts. Recall, Section \ref{sec:leverage_aux_parts} explains how these auxiliary parts are leveraged. 

\begin{algorithm}
{\fontsize{8}{9}\selectfont  
\caption{Auxiliary Part Clustering}
\label{alg:clustering}
\begin{algorithmic}[1]

\Function{GreedyClustering}{$k$,$\alpha$,$\beta$,$\gamma$}

\State $\triangleright$ Initialization
\State $P = \{1,\cdots,n\}$ \Comment{Unclustered patches}
\State $S = \{1,\cdots,n\}$ \Comment{Unclustered seeds}
\State $C = \emptyset$ \Comment{Clusterings initially empty}

\While{$S \; != \emptyset$}
\State{Let $i$ be a random element of $S$} 
\State{$C^* = $ getCluster(i,P,$k$,$\beta$,$\gamma$)}
\If{$|C^*| >= \alpha$} 
\State{$P = P \setminus C^*$} 
\State{$S = S \setminus C^*$} 
\State{$C = C \bigcup \{C^*\}$} \Comment{Add cluster to clusterings}
\Else
\State{$S = S \setminus i$} \Comment{Bad seed patch}
\EndIf
\EndWhile
\State{\textbf{return} $C$}

\EndFunction

\Function{getCluster}{i,P,$k$,$\beta$,$\gamma$} 
\State{$Q = $ getKMostSimilarPatchesToSeed(P,i,$k$)} 
\State{$E = $ getAlignmentEnergyToSeed(i,Q)}
\State{$H = \{q \in Q : E(q) >= \beta \}$ }
\State{$Q = Q \setminus H$}
\State{$D = $ getPatchesDisagreeWithSeed(i,Q,$\gamma$)}
\State{$Q = Q \setminus D$}
\State{\textbf{return} $\{Q \bigcup i\}$}
\EndFunction
\end{algorithmic}
}
\end{algorithm}


\vspace{-0.5cm}
\section{Experiments}
\label{sec:experiments}

All experiments are performed on the hawkmoth dataset described in Section \ref{sec:dataset}. To facilitate our machine-learning based approach we randomly split the 421 annotated images into a training set (211) and testing set (210). All results are based on evaluation on the testing set. 

Pose estimation performance on a given test image is measured by localization error for each semantic part $S_i$. In particular, for  semantic part $S_i$ the localization error is measured by the Euclidean distance between the algorithm's part localization $L_i = (x_i,y_i)$ and human annotated ground truth $L^*_i = (x^*_i,y^*_i)$.

We performed three experiments, each evaluating the accuracy of different algorithms on the hawkmoth dataset. The first experiment establishes a baseline level of performance by applying the MPS model. The second experiment determines the performance gained when using auxiliary parts to update the appearance likelihoods in the MPS baseline. The third experiment establishes how the existing approach of Ortega-Jimenez et al. \cite{Ortega-Jimenez2014} performs on this dataset. 

\begin{figure}[t]
\begin{center}
   \includegraphics[width=1.05\linewidth]{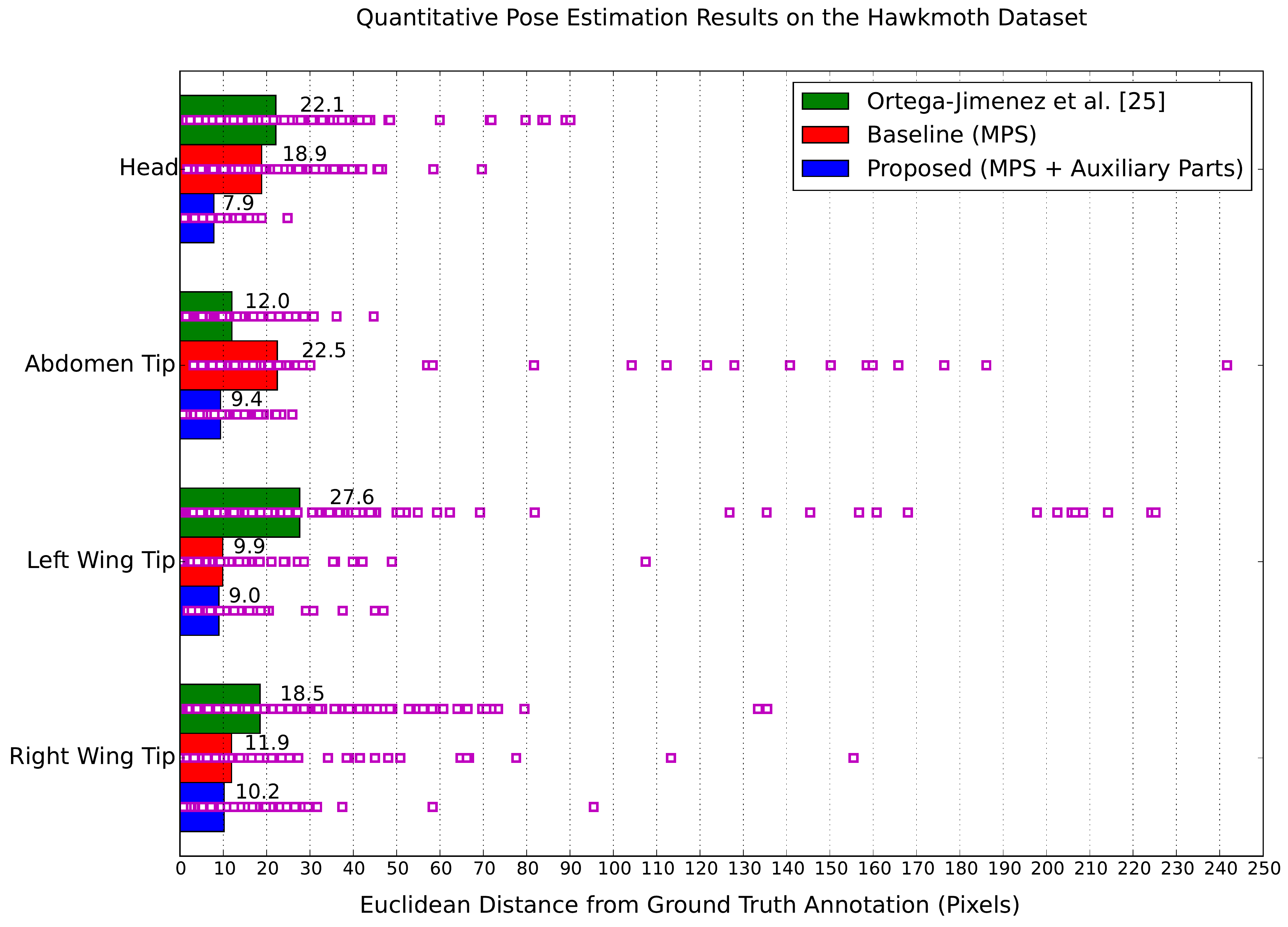} \\
\end{center}
	\vspace{-0.3cm}
   \caption{Quantitative results which summarize error distributions of the baseline, proposed approach,  and Ortega-Jimenez et al. \cite{Ortega-Jimenez2014}, on a hawkmoth test set of 210 images.}
\label{fig:obp}

\end{figure}

Quantitative results for each algorithm are summarized in Figure \ref{fig:obp} and Table \ref{tab:results}. Specifically, Figure \ref{fig:obp} gives a more visual representation of the distribution of errors (purple/magenta squares) for each algorithm on each semantic part. The mean errors are represented by the width of the bar graphs with the numeric value also displayed just to the right of the bar. To help compare the overall distribution of errors across algorithms, Table \ref{tab:results} gives the mean, standard deviation, and the mean squared error (MSE). 

\begin{figure*}[t]
\begin{center}
   \includegraphics[width=0.75\linewidth]{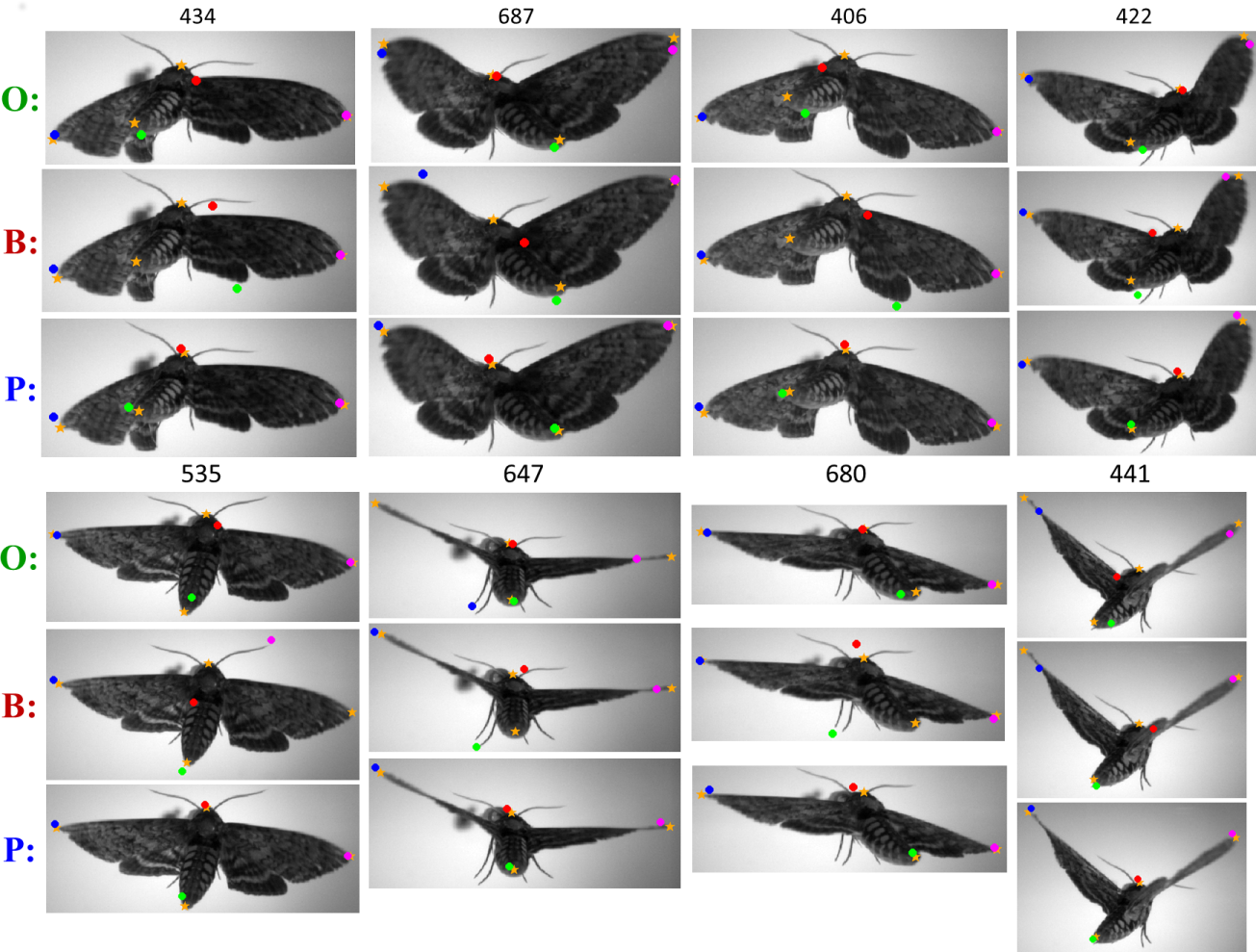} \\
\end{center}
   \caption{Qualitative results for the \textbf{B}aseline, \textbf{P}roposed approach,  and \textbf{O}rtega-Jimenez et al. \cite{Ortega-Jimenez2014} on 8 test images. Orange stars are ground truth annotations. Circles represent part localizations output by the corresponding algorithm. Red, Green, Blue, and Magenta for the Head, Abdomen Tip, Left Wing Tip, and Right Wing Tip respectively.}
\label{fig:qual_results}
\end{figure*}

\begin{figure}[]
\begin{center}
   \includegraphics[width=0.9\linewidth]{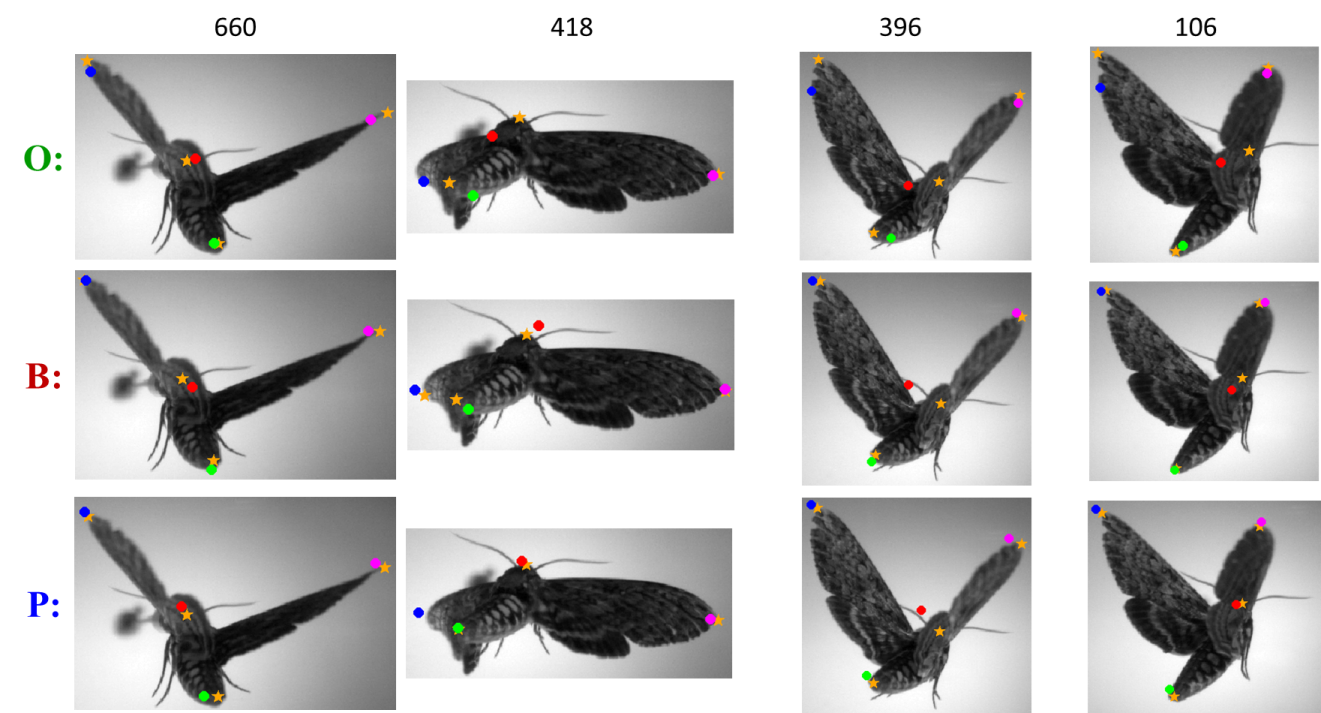} \\
\end{center}
   \caption{Qualitative results for the \textbf{B}aseline, \textbf{P}roposed approach,  and \textbf{O}rtega-Jimenez et al. \cite{Ortega-Jimenez2014} on test images where a semantic part is occluded. The head is occluded by the left wing in frame 660, and by the right wing in frames 396 and 106. In frame 418 the left wing tip is occluded due to its deformation.}
\label{fig:qual_results_occluded}
\end{figure}

\begin{figure}[]
\begin{center}
   \includegraphics[width=0.6\linewidth]{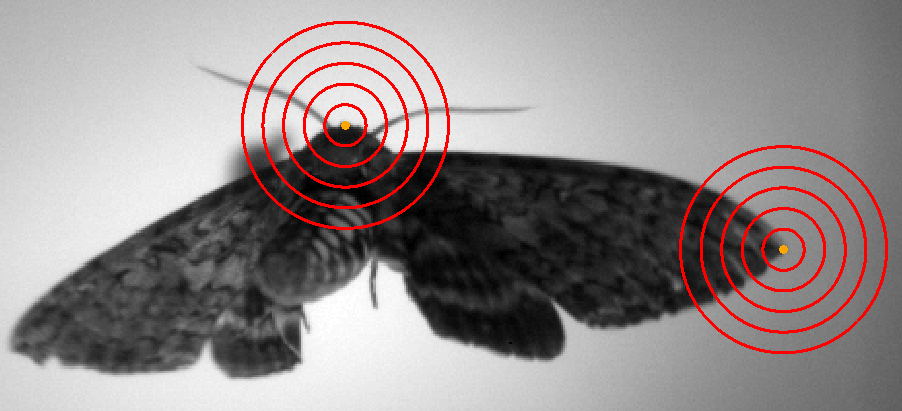} \\
\end{center}
   \caption{Visualization of different levels of localization error. Red rings are drawn with radiuses increasing from 10 pixels to 50 pixels showing how that much localization error looks like relative to the ground truth annotations for the head and right wing tip (orange circles).}
\label{fig:qual_error_vis}
\end{figure}

Qualitative results are shown on 8 test images in Figure \ref{fig:qual_results}, and 4 test images containing occlusions in Figure \ref{fig:qual_results_occluded}. The localizations output automatically by each algorithm are shown as colored circles and ground truth annotations are shown as orange stars. Figure \ref{fig:qual_error_vis} helps connect quantitative error to qualitative error by visualizing what localizations that are 10, 20, 30, 40, and 50 pixels from ground truth look like.

%
%
%

\setlength{\tabcolsep}{3pt}
\begin{table}[t]
\footnotesize
\centering
\caption{Summary of quantitative experimental results. For each semantic part type, the mean $\mu$, standard deviation $\sigma$, and Mean Squared Error (MSE) of the error distribution associated with each algorithm (\textbf{O}rtega-Jimenez, \textbf{B}aseline, and \textbf{P}roposed). Note: all values are rounded.}
\label{tab:results}
\vspace{.2cm}
\begin{tabular}{|c|c|c|c|c|c|c|c|c|c|c|c|c|}

\hline
Alg. & \multicolumn{3}{|c|}{H} & \multicolumn{3}{|c|}{AT} & \multicolumn{3}{|c|}{LWT} & \multicolumn{3}{|c|}{RWT} \\
\hline

$\;$ & $\mu_h$ & $\sigma_h$ & \tiny {\it MSE}$_h$ & $\mu_{a}$ & $\sigma_{a}$ & \tiny {\it MSE}$_{a}$ & $\mu_{l}$ & $\sigma_{l}$ & \tiny {\it MSE}$_{l}$ & $\mu_{r}$ & $\sigma_{r}$ & \tiny {\it MSE}$_{r}$ \\
\hline

\multirow{2}{*}{O} & \multirow{2}{*}{22} & \multirow{2}{*}{17} & \multirow{2}{*}{765} & \multirow{2}{*}{12} & \multirow{2}{*}{8} & 
\multirow{2}{*}{201} & \multirow{2}{*}{28} & \multirow{2}{*}{46} & \multirow{2}{*}{2856} & \multirow{2}{*}{19} & \multirow{2}{*}{21} 
& \multirow{2}{*}{777} \\  
 &  &  &  &  &  &  &  &  &  &  & &  \\
\hline

\multirow{2}{*}{B} & \multirow{2}{*}{19} & \multirow{2}{*}{11} & \multirow{2}{*}{478} & \multirow{2}{*}{23} & \multirow{2}{*}{36} & 
\multirow{2}{*}{1783} & \multirow{2}{*}{10} & \multirow{2}{*}{10} & \multirow{2}{*}{191} & \multirow{2}{*}{12} & \multirow{2}{*}{17} & 
\multirow{2}{*}{419} \\  
&  &  &  &  &  &  &  &  &  &  & &  \\
\hline

\multirow{2}{*}{P} & \multirow{2}{*}{8} & \multirow{2}{*}{3} & \multirow{2}{*}{72} & \multirow{2}{*}{9} & \multirow{2}{*}{4} & \multirow{2}{*}{106} & \multirow{2}{*}{9} & \multirow{2}{*}{6} & \multirow{2}{*}{115} & \multirow{2}{*}{10} & \multirow{2}{*}{9} & \multirow{2}{*}{187} \\ 
&  &  &  &  &  &  &  &  &  &  & &  \\
\hline

\end{tabular}
\vspace{-0.3cm}
\end{table}

%

\section{Discussion}

Our experimental results show quantitatively and qualitatively that our proposed algorithm outperforms the MPS baseline and the work of Ortega-Jimenez et al. \cite{Ortega-Jimenez2014} across all semantic parts. Table \ref{tab:results} makes this clear because the MSE of the proposed approach is not only the lowest among algorithms but it has no more than approximately half the MSE of the next best approach.


When comparing the proposed approach with the MPS baseline we gain an insight into how much and where discovered parts are helping. The largest improvements of the proposed approach over the baseline happen for the abdomen tip and the head. We believe the reason for this gap is that our proposed approach is able to discover that there exist antennae and abdomens (discovered parts), and that they are predictive of where the head and abdomen tip are. This process is demonstrated in Figure \ref{fig:proposed_overview}. Furthermore, the head and abdomen tip are not very discriminative due to their lack of texture, which greatly hinders the performance of the baseline. For cases where body parts are occluded, as in Figure \ref{fig:qual_results_occluded}, both approaches are able to guess where the occluded part should be located. This can be attributed to the spatial terms of the PS models that learn common configurations of body parts. The proposed approach is also potentially advantageous in occlusion cases as is demonstrated by the fact that antennae can help predict an occluded head.

The MPS baseline is a baseline we created to represent works that extend pictorial structures both with global mixtures and local (part-level) mixtures. We feel our comparison with this baseline accurately reflects the advantage our proposed approach over these types of part-based models. 

One of the core challenges in our work involved discovering parts from the hawkmoth dataset. In practice this meant discovering which features and clustering algorithms would work well for this application. We found that dense SIFT was useful in capturing the finer details of the hawkmoth's texture. For clustering patches, which are not aligned in anyway apriori, we found it important to use the alignment energy computed with SIFT Flow\cite{Liu2011} as a way to remove outliers from clusters.    


\section{Conclusion}

Our work introduces a novel way to increase pose estimation accuracy by using automatically discovered auxiliary parts to generate better appearance likelihoods which can then be fed into traditional part-based models like the MPS model. Our experiments on the hawkmoth dataset give quantitative and qualitative support to the value of our proposed approach over traditional part-based models. Furthermore, our approach yields significantly more accurate hawkmoth part localizations than previous work \cite{Ortega-Jimenez2014} while being more general in applicability.

We hope our proposed approach will inspire more works to think about ways to leverage unannotated regions of training images for pose estimation / landmark localization problems. We also think it is important for biology datasets to get more attention from the mainstream computer vision community. To facilitate both of these aims we are making our unique hawkmoth dataset along with annotations and segmentations publicly available. Future work entails extending our proposed approach to a multi-view dataset to obtain more accurate analyses of 3D hawkmoth flight.

\small \textbf{Acknowledgments}: This work was partially funded by ONR (N000141010952) and NSF (0910908, 1253276).

{\small
\bibliographystyle{ieee}
\bibliography{./moth_paper}
}

\end{document}